%% file: main.tex
\documentclass[10pt]{article}
\usepackage[preprint]{ivgpaper}

\input{paper_preamble.tex}
\input{paper_metadata.tex}

\begin{document}
\maketitle

\input{sections/abstract}
\input{sections/teaser}
\input{sections/introduction}
\input{sections/related_work}
\input{sections/method}
\input{sections/experiments}
\input{sections/conclusion}

\bibliographystyle{plainnat}
\bibliography{references}

\clearpage
\appendix
\input{sections/appendix}

\end{document}

%% file: paper_preamble.tex
\usepackage{graphicx}
\usepackage{booktabs}
\usepackage{multirow}
\usepackage{array}
\usepackage{tabularx}
\usepackage{makecell}
\usepackage{adjustbox}
\usepackage{wrapfig}
\usepackage{colortbl}
\usepackage{subcaption}
\usepackage{float}
\usepackage{placeins}
\usepackage{amsmath}
\usepackage{amssymb}
\usepackage{bm}
\usepackage{algorithm}
\usepackage{algorithmic}
\usepackage[numbers,sort&compress]{natbib}
\usepackage{url}
\usepackage{pifont}
\usepackage[normalem]{ulem}
\usepackage{tikz}
\usetikzlibrary{arrows.meta,positioning,fit,calc,shapes.geometric,backgrounds}

\newcommand{\cmark}{\ding{51}}
\newcommand{\xmark}{\ding{55}}

\input{math_commands.tex}

\newcommand{\shortname}{GigaBrain-WBC-0.5}

\newcommand{\refwin}{\bm{c}}          
\newcommand{\lat}{\bm{z}}             
\newcommand{\latraw}{\lat^{\mathrm{raw}}}
\newcommand{\latfin}{\lat^{\star}}
\newcommand{\prop}{\bm{s}}            
\newcommand{\act}{\bm{a}}             
\newcommand{\gmm}{\mathcal{G}}
\newcommand{\Mah}{M^{2}}
\newcommand{\Rsafe}{R_{\mathrm{safe}}}
\newcommand{\Rsq}{\Rsafe^{2}}

\usepackage[colorlinks=true,linkcolor=ivgblue,citecolor=ivgblue,urlcolor=ivgblue]{hyperref}
\usepackage[capitalize,nameinlink,noabbrev]{cleveref}

%% file: math_commands.tex
\usepackage{amsmath,amsfonts,bm}

\def\eqref#1{equation~\ref{#1}}

\def\1{\bm{1}}

\DeclareMathAlphabet{\mathsfit}{\encodingdefault}{\sfdefault}{m}{sl}
\SetMathAlphabet{\mathsfit}{bold}{\encodingdefault}{\sfdefault}{bx}{n}



%% file: paper_metadata.tex
\ivgtitle{GigaBrain-WBC-0.5: A Behavior World Model\\ for Robust Humanoid Whole-Body Tracking\\ with Environment Interaction}
\ivgrunningtitle{GigaBrain-WBC-0.5}
\ivgvenue{GigaBrain-WBC-0.5 Technical Report}
\ivgsetlogo{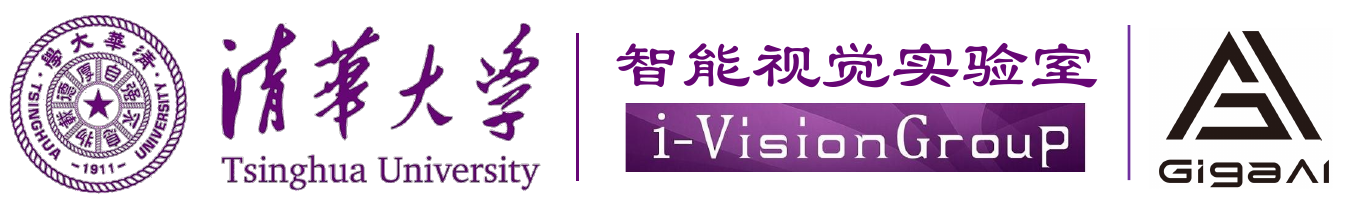}

\ivgauthor{%
  {\small
  Ziyang Cheng\textsuperscript{1,2}\hspace{0.45em}
  Tianshu Tang\textsuperscript{1,2}\hspace{0.45em}
  Jinxin Lan\textsuperscript{1,2}\hspace{0.45em}
  Xinze Chen\textsuperscript{2}\hspace{0.45em}
  Yuhan Gong\textsuperscript{2} \\
  Zhichao Liu\textsuperscript{2}\hspace{0.45em}
  Changzhong Wu\textsuperscript{2}\hspace{0.45em}
  Yahao Mao\textsuperscript{2,3}\hspace{0.45em}
  Zongyan Deng\textsuperscript{2,4}\hspace{0.45em}
  Mingxuan Ma\textsuperscript{2,4} \\
  Huasen Xi\textsuperscript{2,5,6}\hspace{0.45em}
  Yilong Liu\textsuperscript{1,2}\hspace{0.45em}
  Yutong Wu\textsuperscript{2}\hspace{0.45em}
  Xiaofeng Wang\textsuperscript{1} \\
  Borui Zhang\textsuperscript{1}\hspace{0.45em}
  Bingyao Yu\textsuperscript{1}\hspace{0.45em}
  Yang Wang\textsuperscript{2} \\
  Yun Ye\textsuperscript{2}\hspace{0.45em}
  Guan Huang\textsuperscript{2}\hspace{0.45em}
  Xiaojie Jin\textsuperscript{4}\hspace{0.45em}
  Zheng Zhu\textsuperscript{2,\#}\hspace{0.45em}
  Jiwen Lu\textsuperscript{1,\#}\\[0.45em]
  }
  {\footnotesize
  \textsuperscript{1}Tsinghua University, \hspace{0.45em}
  \textsuperscript{2}GigaAI, \hspace{0.45em}
  \textsuperscript{3}University of Shanghai for Science and Technology\\
  \textsuperscript{4}Beijing Jiaotong University, \hspace{0.45em}
  \textsuperscript{5}Institute of Automation, Chinese Academy of Sciences\\
  \textsuperscript{6}University of Chinese Academy of Sciences\\[0.25em]
  }
  {\footnotesize
  \textsuperscript{\#}Corresponding authors.}\\[0.25em]
  {\footnotesize Project page: \href{https://shepherd1226.github.io/gigabrain-wbc-0.5/}{https://shepherd1226.github.io/gigabrain-wbc-0.5/}}
}

%% file: sections/abstract.tex
\begin{abstract}
General-purpose motion trackers enable humanoid robots to follow diverse whole-body motions while maintaining balance, but are trained only on flat ground, failing to exploit bipedal mobility over complex terrain.
Cross-terrain controllers, meanwhile, are task-specific or accept only low-dimensional locomotion commands.
We introduce \shortname{}, the first behavior world model (BWM) for robust whole-body tracking with environment interaction.
Its Transformer jointly predicts the next action, state, and behavior distribution, learning environment-conditioned dynamics.
To scale interaction training data, an automatic annotation pipeline reconstructs 3D support geometry from retargeted motions.
At deployment, the policy handles commands implausible in the current environment in a ``best-effort'' manner.
Quantitatively, \shortname{} achieves an $81.3\%$ success rate on terrain interaction ($4.3\times$ the best evaluated baseline) and a $99.3\%$ fall-recovery rate, while also improving free-space tracking and outperforming three leading tracking baselines across all of these regimes.
To our knowledge, we provide the first demonstration of real-time cross-terrain whole-body teleoperation on a humanoid robot, alongside object interaction, stable responses to missing supports, and robust recovery from falls.
\end{abstract}

%% file: sections/teaser.tex
\begin{figure}[!t]
  \centering
  \includegraphics[width=\linewidth]{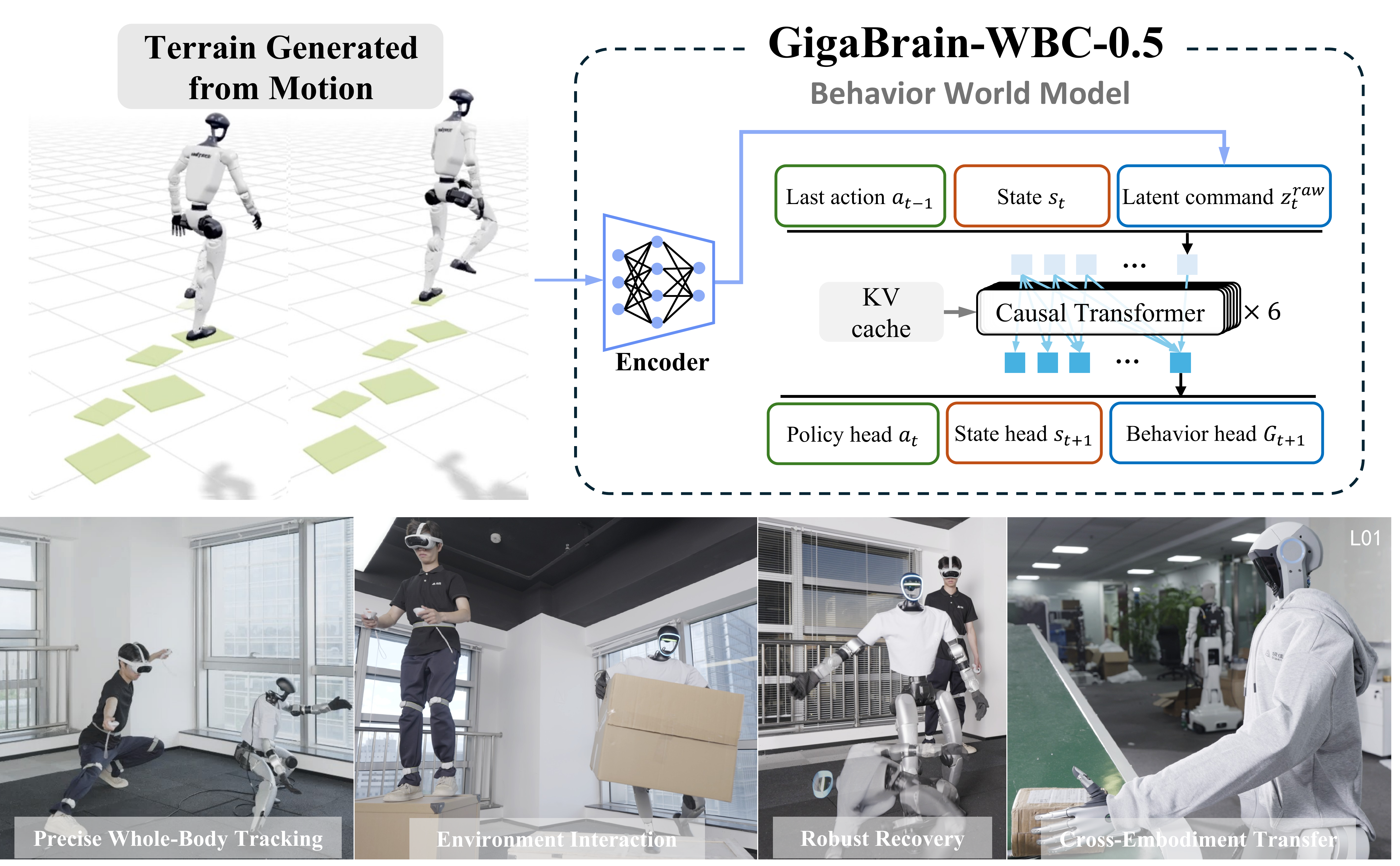}
  \caption{\textbf{\shortname{} learns environment-conditioned whole-body tracking.}
  A behavior world model predicts future states and behaviors, thereby capturing environment-conditioned dynamics; the resulting unified policy supports precise tracking, environment interaction, and fall recovery under live whole-body commands. Fine-tuning transfers the Unitree G1 checkpoint to Maker L01.}
  \label{fig:teaser}
\end{figure}

%% file: sections/introduction.tex
\section{Introduction}
\label{sec:intro}

General-purpose motion tracking turns a humanoid into a general control interface.
A single policy can imitate diverse human motions while handling low-level balance and physical feasibility, allowing a human operator \citep{ze2025twist,he2024omnih2o}, planner, or vision-language-action model \citep{bjorck2025gr00t} to provide coarse motion intent instead of precise torque-level commands.
Recent trackers scale this recipe to hundreds or thousands of hours of retargeted motion \citep{luo2025sonic,chen2026holomotion,qi2026humanoidgpt,chen2025gmt,zhang2025track,liao2025beyondmimic}, enabling agile teleoperation and data collection \citep{ze2025twist2}.

The advantage of humanoid legs lies in traversing and interacting with complex environments, yet existing trackers are typically trained only on flat ground in empty scenes.
A general-purpose tracker trained only in this setting therefore never learns that a contact surface can change the robot's dynamics or enable or preclude a commanded behavior: a chair makes it possible to sit down but prevents stepping backward, while a payload shifts the center of mass and requires the policy to adapt in order to maintain balance.
Policies that do support rich interaction or locomotion across varied terrain are either specialized for particular skills \citep{yang2025omniretarget,weng2025hdmi,zhao2025resmimic,zhang2024wococo} or limited to directional velocity commands for locomotion, precluding general manipulation tasks \citep{zhang2026rpl,ben2025gallant,zhu2026hiking,zhang2026ame2}.
Training interaction at motion-corpus scale requires pairing each motion with consistent scene geometry, which ordinary motion-capture datasets cannot provide.
SceneBot \citep{chen2026scenebot} reconstructs terrain post hoc but uses 2.5D elevation maps and per-link contact prompts; height fields cannot represent overhangs or vertical supports, and the required contact labels are unavailable from a live operator.

Environment interaction also compounds the robustness problem.
On flat ground, prior work simply broadens the reference corpus until many commands become learned and executable.
With varied geometry, feasibility is conditional: sitting down is meaningful on a chair but not in free space, and a level-ground gait may be viable on the floor but not on stairs.
Moreover, simple fallbacks become inadequate: emergency stops or standing-pose recovery can not only interrupt the task but also create safety risks when the robot is performing a dynamic task on complex terrain.
A useful controller should therefore preserve as much of the command as its current situation permits.
It should also remain controllable after a disturbance or fall, rather than interrupting the task to invoke a separate recovery policy.

We introduce \shortname{}, a \emph{behavior world model} for robust humanoid whole-body tracking with environment interaction.
Its causal Transformer emits actions and predicts the next proprioceptive state and the conditional distribution over the next latent behavior command, learning how the current environment affects robot dynamics and which behaviors it permits.
At deployment, the filter uses this distribution to project implausible commands toward compatible behaviors, producing a stable ``best-effort'' response that retains intent.
Our automatic pipeline reconstructs 3D support geometry from contact evidence in retargeted trajectories, providing paired motion--geometry data for interaction training at motion-corpus scale.
In simulation, \shortname{} reaches $81.3\%$ terrain-interaction success ($4.3\times$ the best baseline) and $99.3\%$ fall recovery, with the best overall performance on standard free-space tracking.
Hardware trials show live whole-body teleoperation across terrain, object interaction, missing-support robustness, and disturbance recovery.

Our contributions are:
\begin{itemize}
  \item a causal behavior world model that jointly predicts actions, future states, and future behavior distributions for environment-conditioned whole-body tracking;
  \item an automatic annotation pipeline that reconstructs 3D support geometry from ordinary retargeted motions, scaling interaction training beyond scene-captured datasets; and
  \item stable ``best-effort'' behavior under implausible commands and recovery from falls within a unified tracking policy.
\end{itemize}

%% file: sections/related_work.tex
\section{Related Work}
\label{sec:related}

\begin{table}[t]
  \caption{\textbf{Capability comparison.} Representative methods.}
  \label{tab:capability}
  \centering
  \small
  \setlength{\tabcolsep}{4.5pt}
  \begin{tabular*}{\linewidth}{@{\extracolsep{\fill}}lcccccc@{}}
    \toprule
    & \makecell{Diverse\\whole-body\\tracking} & \makecell{Low-\\latency\\teleop.} & \makecell{Terrain\\interaction} & \makecell{Object\\interaction} & \makecell{Robust to\\implausible\\commands} & \makecell{Robust\\to falls} \\
    \midrule
    GMT \citep{chen2025gmt}                   & \cmark & \xmark & \xmark & \xmark & \xmark & \xmark \\
    SONIC \citep{luo2025sonic}                & \cmark & \cmark & \xmark & \xmark & \xmark & \xmark \\
    Humanoid-GPT \citep{qi2026humanoidgpt}    & \cmark & \cmark & \xmark & \cmark & \xmark & \xmark \\
    PHP \citep{wu2026php}                     & \xmark & \xmark & \cmark & \xmark & \xmark & \xmark \\
    OmniRetarget \citep{yang2025omniretarget} & \xmark & \xmark & \cmark & \cmark & \xmark & \xmark \\
    CMP \citep{cheng2026cmp}                  & \cmark & \cmark & \xmark & \cmark & \cmark & \xmark \\
    BFM-Zero \citep{li2025bfmzero}            & \cmark & \cmark & \xmark & \xmark & \cmark & \cmark \\
    SceneBot \citep{chen2026scenebot}         & \cmark & \xmark & \cmark & \cmark & \xmark & \xmark \\
    Perceptive BFM \citep{wang2026perceptivebfm}                  & \cmark & \xmark$^{\ddagger}$ & \cmark & \xmark & \xmark & \cmark \\
    PGMT \citep{li2026pgmt}                  & \cmark & \xmark$^{\dagger}$ & \cmark & \xmark & \xmark & \cmark \\
    \midrule
    \textbf{\shortname{} (ours)}              & \cmark & \cmark & \cmark & \cmark & \cmark & \cmark \\
    \bottomrule
  \end{tabular*}
  \parbox{\linewidth}{\footnotesize $^{\dagger}$PGMT postdates this work's first public release; its demo shows over 1\,s of teleoperation delay, exceeding our low-latency criterion. $^{\ddagger}$Perceptive BFM's 21-frame future window implies a look-ahead delay of at least $0.4$\,s at 50\,Hz.}
\end{table}

\paragraph{Whole-body tracking and behavior foundation models.}
Physics-based imitation \citep{peng2018deepmimic,peng2021amp} made motion capture a scalable source of dense control supervision.
Modern systems expand motion diversity with quantized latent commands \citep{luo2025sonic}, sparse mixture-of-experts Transformers \citep{chen2026holomotion}, expert distillation \citep{qi2026humanoidgpt}, diffusion or masked completion \citep{liao2025beyondmimic,tessler2024maskedmimic}, and heterogeneous command interfaces \citep{yin2025unitracker,wang2025bumblebee,he2025hover}.
Other work emphasizes agile skills \citep{he2025asap,han2025kungfubot2}, perturbation robustness \citep{zhang2025track}, or the fidelity--diversity trade-off \citep{wang2026omnixtreme,luo2023universal}.
Teleoperation systems \citep{ze2025twist,ze2025twist2,he2024omnih2o,ben2025homie} expose these policies as real-time interfaces.
Behavior foundation models extend shared pretraining across control modes \citep{yuan2025bfmsurvey,zeng2025bfm,li2025bfmzero}.
Most whole-body trackers are restricted to flat ground.
The few exceptions---Perceptive BFM \citep{wang2026perceptivebfm}, PGMT \citep{li2026pgmt}, and SceneBot \citep{chen2026scenebot}---respectively require at least $0.4$\,s of future-reference buffering, exhibit over $1$\,s of teleoperation delay in demonstrations, and rely on per-link contact prompts unavailable from live motion capture.
\shortname{} scales up interaction data approximately tenfold while supporting real-time cross-terrain whole-body teleoperation.

\paragraph{Interaction with terrain and objects.}
Classical contact planning and model-predictive control achieve structured locomotion \citep{deits2014footstep,kuindersma2016atlas}; learned perceptive policies extend this to stairs, slopes, constrained terrain, and parkour \citep{zhang2026rpl,ben2025gallant,zhu2026hiking,zhang2026ame2,wu2026php}.
Scene-aware retargeting and data generation support large-object or whole-body interaction \citep{yang2025omniretarget,weng2025hdmi,zhao2025resmimic,zhang2024wococo,jiang2022tip}, while video reconstruction and terrain-aware generation couple motions to geometry \citep{zhang2026meshmimic,xu2025parc,zhang2026learning}.
These systems are skill-specific or accept only directional velocity commands, providing joystick-level locomotion insufficient for complex manipulation.
\shortname{} instead controls terrain and object interaction from whole-body motion commands.

\paragraph{Robustness to commands and falls.}
Domain randomization and other perturbations widen a tracker's training distribution \citep{zhang2025track,chen2025chip,he2025asap}, while learned priors constrain behavior to a motion manifold \citep{luo2023universal,tessler2024maskedmimic}.
In legged manipulation, CMP \citep{cheng2026cmp} shows that long-horizon command safety reduces to a per-step test and that the distributions of training and feasible commands closely match.
We extend both insights to humanoid whole-body tracking, where command feasibility also depends on terrain and whole-body contacts.
BFM-Zero further shows that a well-regularized latent admits graceful degradation under out-of-distribution prompts.
Fall recovery is typically delegated to specialized standing or protective policies \citep{huang2025host,meng2025safefall}; BFM-Zero \citep{li2025bfmzero} shows that it can be integrated into a general policy.
\shortname{} further unifies fall recovery, environment interaction, and agile teleoperation in one tracker that resumes live commands without mode switching.

%% file: sections/method.tex
\section{Method}
\label{sec:method}

\begin{figure}[t]
  \centering
  \includegraphics[width=\linewidth]{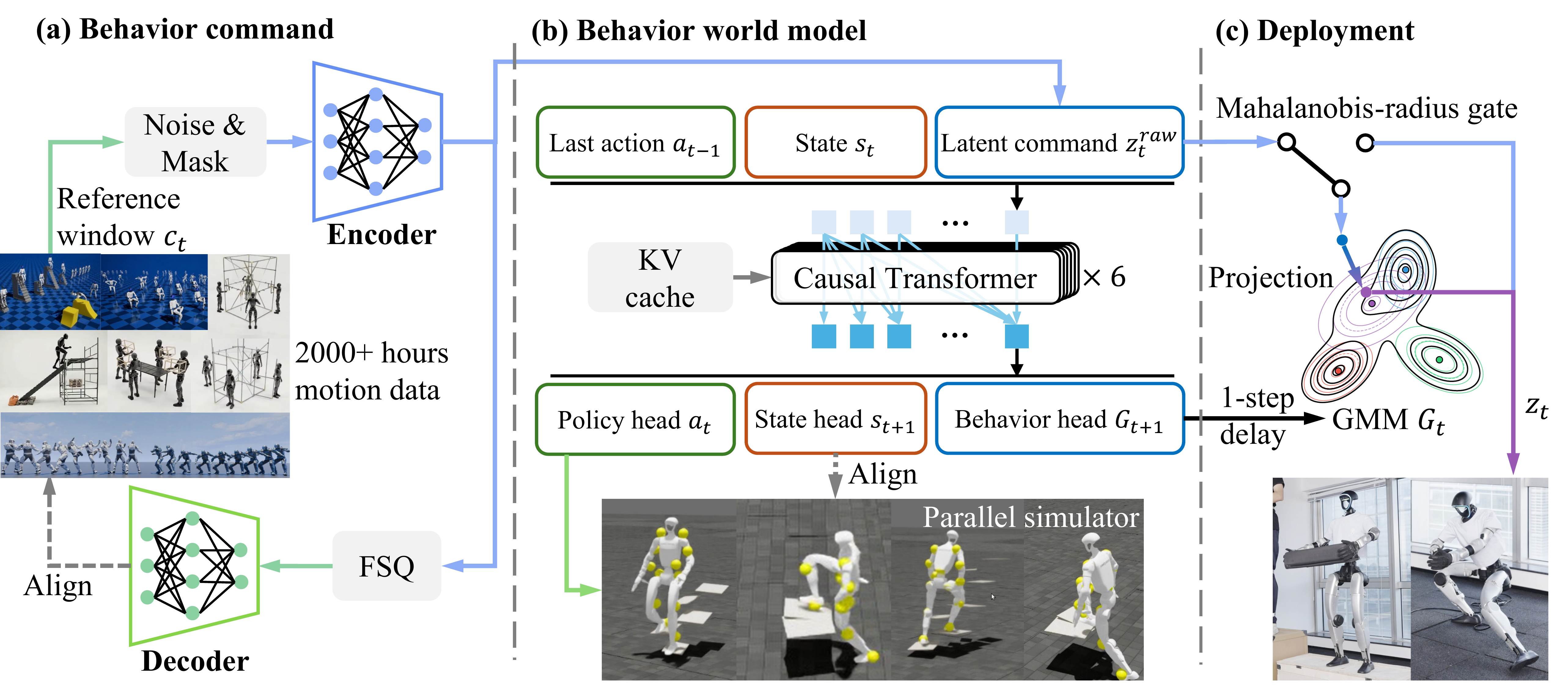}
  \caption{\textbf{\shortname{} overview.}
  A noisy, partially masked reference window is encoded into a continuous behavior command; a finite scalar quantizer (FSQ) regularizes this latent through a training-only reconstruction branch.
  The causal Transformer consumes the command, proprioception, and previous action, then jointly predicts the action, next state, and next-command distribution.
  At deployment, the preceding distribution identifies and corrects implausible commands before control.}
  \label{fig:overview}
\end{figure}

\subsection{Problem Setting}

We control a Unitree G1 through joint-level PD targets.
At step $t$, the policy maps proprioception $\prop_t$, the previous target $\act_{t-1}$, and a short reference window $\refwin_t$ to the next target $\act_t$; \cref{app:input} specifies the exact interface.
The reference comes from teleoperation or a high-level policy.

Like other general-purpose trackers, ours deliberately omits vision, following a brain--cerebellum division of labor: an upstream system perceives the scene and selects the reference, while the compact, high-rate controller focuses on dynamics.
Uncontacted geometry cannot affect the current dynamics; once contact occurs, its effect is exposed more directly through proprioception than vision.
Thus, when terrain and objects make execution depend on supports, contacts, and loads, the controller infers these conditional dynamics from proprioceptive history without scene observations or privileged contact labels.

\subsection{Behavior World Model}
\label{sec:bwm}

To make the controller understand and adapt to environment-conditioned dynamics, we train it as a behavior world model rather than an action predictor alone (\cref{fig:overview}).
Alongside the action, it predicts the robot's next proprioceptive state and the conditional distribution of the next behavior command.
Predicting the next state teaches the controller how current contacts change the robot's dynamics, while predicting the next command teaches it which behaviors remain compatible with the current physical situation.
Together, these objectives let the controller account for interaction dynamics and behavior feasibility when choosing an action.

A reference encoder first maps the reference window $\refwin_t$ to a continuous raw behavior command $\latraw_t\in\mathbb R^{d_z}$, where $d_z=64$; its unchanged or corrected version $\lat_t$ conditions the controller.
The core input $\bm e_t=[\prop_t,\act_{t-1},\lat_t]$ is then processed by a six-layer causal Transformer \citep{vaswani2017attention} with rotary embeddings \citep{su2024roformer}.
Each layer attends over 32 frames, yielding a 187-frame effective receptive field that captures long-horizon interaction history.
A per-layer KV cache accelerates autoregressive inference.
From the Transformer output at step $t$, three heads produce
\begin{equation}
  \act_t\in\mathbb{R}^{29},\qquad
  \hat{\prop}_{t+1}\in\mathbb{R}^{67},\qquad
  \gmm_{t+1}=\{\pi_k,\bm\mu_k,\log\bm\sigma_k\}_{k=1}^{4},
  \label{eq:heads}
\end{equation}
where $\pi_k$ are the predicted mixture weights and $\gmm_{t+1}(\lat)=\sum_{k=1}^{4}\pi_k\mathcal N(\lat;\bm\mu_k,\operatorname{diag}(\bm\sigma_k^2))$.
This standard diagonal mixture allows distinct behavior modes to be represented without conflating them in one broad Gaussian.

Because both the command predictor and deployment filter operate in the latent space, its geometry directly determines whether the mixture can represent and delimit distinct behavior modes.
We therefore attach a training-only FSQ \citep{mentzer2023finite} branch to the reference encoder.
As in SONIC \citep{luo2025sonic}, this auxiliary branch decodes the quantized command back to the clean reference and re-encodes the reconstruction for cycle-consistency supervision.
Without quantization, minimizing the cycle-consistency MSE can favor a shrinking latent scale, pulling encoded behaviors progressively closer together.
FSQ prevents this shortcut: if the scale collapses, nearby behaviors fall into the same quantization bins and can no longer retain enough information for reconstruction, while the finite number of levels also limits how much information can be folded into any one latent dimension.
Unlike SONIC, however, our main Transformer branch bypasses FSQ and always consumes the continuous command, preserving fine motion variations at deployment instead of exposing control to discrete token transitions.

The reference encoder must also reconcile long-horizon terrain alignment with hardware observability.
Our 187-frame receptive field calls for long episodes: frequently resetting them to bound global drift would truncate the history the model is designed to exploit, whereas leaving drift uncorrected eventually misaligns the reference with its paired terrain.
The robot-relative root-translation input provides this correction but requires a global root position unavailable on hardware.
We therefore expose it intermittently during training, teaching the policy to re-anchor when it is present and rely on interaction history when it is absent; it remains masked throughout deployment.

The PPO \citep{schulman2017proximal} objective is
\begin{equation}
\mathcal L=\mathcal L_{\mathrm{PPO}}
+\lambda_{\mathrm{recon}}\mathcal L_{\mathrm{recon}}
+\lambda_{\mathrm{cycle}}\mathcal L_{\mathrm{cycle}}
+\lambda_{\mathrm{state}}\lVert\hat{\prop}_{t+1}-\prop_{t+1}\rVert^2
-\lambda_{\mathrm{beh}}d_z^{-1}\log\gmm_{t+1}(\lat_{t+1}),
\label{eq:objective}
\end{equation}
Loss weights, optimization, reward, curriculum, and randomization details are in \cref{app:training}.

\subsection{Automatic Spatial Terrain Annotation}
\label{sec:terrain}

\begin{figure}[t]
  \centering
  \includegraphics[width=\linewidth]{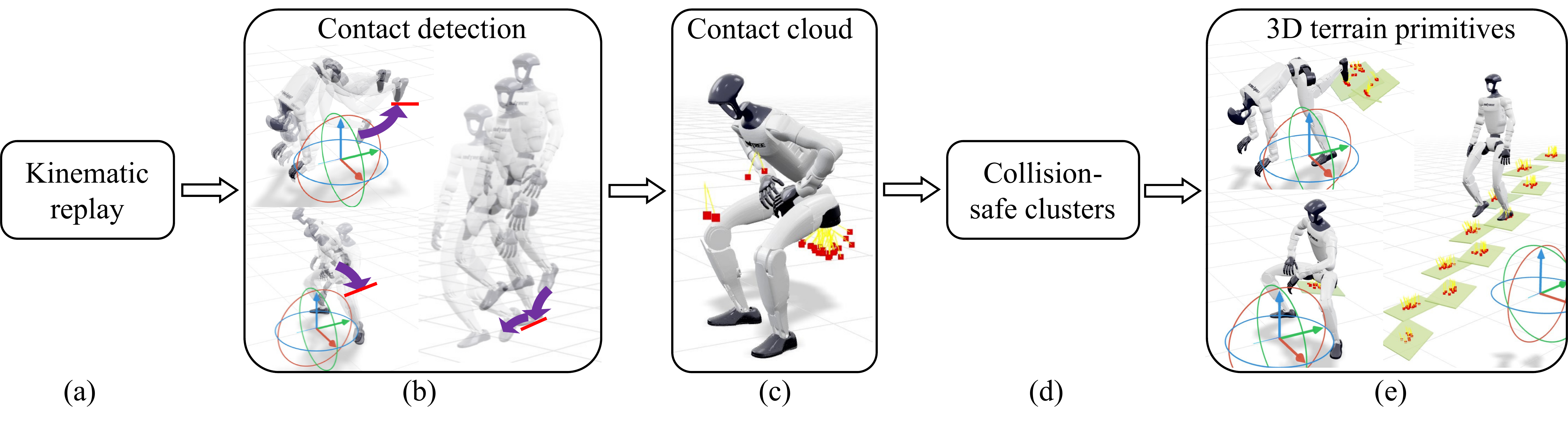}
  \caption{\textbf{Motion-to-terrain annotation.}
  Surface samples that exhibit a contact deceleration are accumulated, filtered against whole-body penetration, clustered, and fitted with spatial collision primitives.}
  \label{fig:terrain}
\end{figure}

Contact-conditioned training requires motion--geometry pairs, but existing routes scale poorly.
MeshMimic \citep{zhang2026meshmimic} reconstructs motions and scenes from video, inheriting video-quality limits;
OmniRetarget \citep{yang2025omniretarget} relies on prebuilt scenes, leaving diversity asset-bound.
We instead infer support geometry from retargeted trajectories (\cref{fig:terrain}), utilizing the large scale of existing motion-capture corpora.
SceneBot \citep{chen2026scenebot} takes a similar approach but reconstructs 2.5D height fields, whereas our full 3D geometry can capture free space beneath structures such as tables.

For each retargeted trajectory, we uniformly sample the collision surfaces of contact-relevant links and replay the motion kinematically in MuJoCo \citep{todorov2012mujoco}, recording world-space positions and outward normals.
A Savitzky--Golay filter \citep{savitzky1964smoothing} suppresses numerical jitter before a per-sample state machine detects contact:
low normal and tangential speeds must coincide with recent sharp normal deceleration, distinguishing support from momentarily slow motion.
Contact ends upon excessive displacement or sliding.
Detected contacts are accumulated into a point cloud.

\begin{wrapfigure}[14]{r}{0.42\textwidth}
  \centering
  \vspace{-14pt}
  \includegraphics[width=\linewidth]{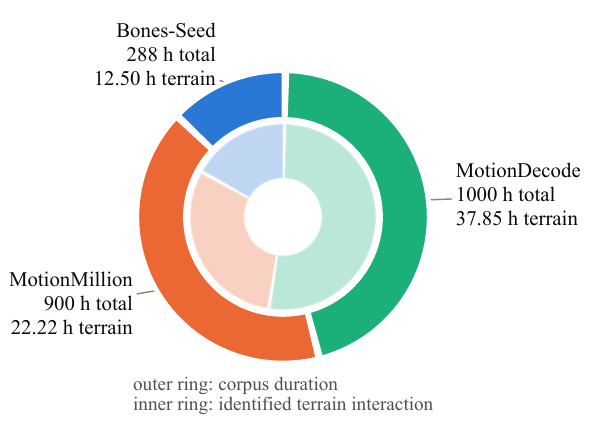}
  \vspace{-15pt}
  \caption{\textbf{Source corpora and terrain coverage.}
  Outer and inner rings show total duration and identified terrain-interaction duration, respectively.}
  \label{fig:datasets}
  \vspace{-6pt}
\end{wrapfigure}
We discard candidates intersecting the robot at any replay frame, preventing unwanted collisions.
An affinity graph links survivors with aligned normals and small normal-direction offsets.
Density-based spatial clustering of applications with noise (DBSCAN) \citep{ester1996density} separates distinct supports, including adjacent surfaces at different heights.
Tangent-plane principal component analysis (PCA) estimates each cluster's principal directions and uses geometric primitives to approximate the point cloud.
Clusters whose bounding boxes contain large empty regions are recursively split along the principal axis to fit non-convex or annular structures.
Finally, a penetration-safe binary search expands each primitive laterally to tolerate tracking errors.

We apply the pipeline to three retargeted corpora (\cref{fig:datasets}): Bones-Seed \citep{bonesseed} (288\,h), MotionMillion \citep{fan2025motionmillion} (900\,h), and MotionDecode \citep{motiondecode} (1000\,h), identifying 12.50, 22.22, and 37.85 hours of terrain interaction.
The resulting terrain primitives are instantiated as static bodies in Isaac Lab \citep{mittal2025isaaclab} and paired with their corresponding reference motions during training.
We evaluate the accuracy of the resulting annotations in \cref{app:terrain_audit}.

\subsection{Robustness to Implausible Commands and Falls}
\label{sec:filter}

Command feasibility depends on the environment: a command that is valid on certain terrain may be physically implausible on flat ground, making blind execution unsafe. Simple solutions, such as an emergency stop or recovery to a standing pose, interrupt the task and may themselves be unsafe while the robot is traversing complex terrain.
Following CMP \citep{cheng2026cmp}, we use a per-step command-distribution test as a practical proxy for command feasibility. With paired terrain--motion training data, we flag commands that are out of distribution (OOD) under the behavior distribution conditioned on the current environment, even if the same motion appears elsewhere in the dataset.
The preceding prediction $\gmm_t$, formed before observing $\refwin_t$, defines this test from the current physical history (\cref{fig:overview}):
$\refwin_t\!\rightarrow E\!\rightarrow\latraw_t\xrightarrow{\gmm_t}\latfin_t\!\rightarrow\mathrm{BWM}\!\rightarrow(\act_t,\hat\prop_{t+1},\gmm_{t+1})$; the first step after initialization or reset is unfiltered.

The mixture components represent distinct behavior modes. For example, from standing, walking commands at different speeds may occupy one region and jumps of different heights another.
We first identify which behavior mode the user most likely intends.
Specifically, we select the mode with maximum posterior responsibility under the standard GMM:
\begin{equation}
k^\star=\arg\max_k\left[\log\pi_k-\frac{1}{2}\sum_{i=1}^{d_z}\left(\left(\frac{z_i-\mu_{k,i}}{\sigma_{k,i}}\right)^2+2\log\sigma_{k,i}+\log2\pi\right)\right].
\label{eq:component}
\end{equation}
Let $\Mah_k(\lat)=\sum_{i=1}^{d_z}((z_i-\mu_{k,i})/\sigma_{k,i})^2$ denote the squared Mahalanobis distance to component $k$.
For the selected Gaussian, $\Mah_{k^\star}(\lat)=\Rsq$ is an equal-density contour;
we treat its interior $\Mah_{k^\star}(\lat)\le\Rsq$, where density exceeds this threshold, as the admissible region of behaviors likely feasible in the current environment.
A command inside this region passes unchanged; otherwise, we retract it toward the selected mode:
\begin{equation}
\latfin_t=\bm\mu_{k^\star}+\sqrt{\Rsq/\Mah_{k^\star}(\latraw_t)}\,(\latraw_t-\bm\mu_{k^\star}).
\label{eq:projection}
\end{equation}
This radial rescaling reaches the ellipsoid boundary along the requested direction, retaining as much intent as the selected mode permits.
We use the Mahalanobis radius $\Rsafe=3$ (thus $\Rsq=9$) and analyze its sensitivity in \cref{app:tradeoff}.
We additionally train physical robustness: fallen-state initialization enables fall recovery, while persistent random external forces at the wrists and torso prepare the policy for object interaction.
At deployment, the command filter neither blindly executes an implausible command nor resorts to an emergency stop, but produces a stable best-effort behavior.

%% file: sections/experiments.tex
\section{Experiments}
\label{sec:exp}

We show that \shortname{} acquires interaction capabilities and robustness while also improving free-space tracking performance. We further evaluate the accuracy of the automatic terrain-generation pipeline and conduct ablations of the behavior world model and terrain-annotation modules.

\subsection{Experimental Setup}
\label{sec:exp:setup}

\paragraph{Benchmarks and metrics.}
\textbf{Standard} contains 136 clips from the AMASS \citep{Mahmood_2019_ICCV} test split.
\textbf{Terrain} contains 150 held-out Bones-Seed \citep{bonesseed} interaction clips with their annotated geometry.
\textbf{Implausible} contains 136 MotionMillion \citep{fan2025motionmillion} clips with self-penetration, missing supports, or motions incompatible with the robot morphology.
\textbf{Fall} uses the Standard references but initializes the robot in fallen poses.
Mean per-keypoint position error (MPKPE) measures tracked-link error; RootPos and RootVel measure root position and linear-velocity error; joint jerk measures recovery smoothness.
Standard, Terrain and Implausible SR require tracking without falling.
Fall SR requires the robot to stand up and resume tracking.
All clips run to completion, including failed trials, avoiding survivorship bias in error metrics.
Further evaluation criteria are provided in Appendix~\ref{app:training}.

Metrics are selected by the question posed by each split.
Standard uses root velocity because absolute position on an unstructured floor is arbitrary, whereas Terrain uses root position because a few centimeters of drift can place a foot on the wrong side of a step.
For the Implausible split, tracking-error magnitude is secondary because the references are physically implausible: we report only MPKPE to coarsely measure retained intent, while SR is primary and here means survival without dangerous collisions or falls.
Fall begins in a failed state, making ordinary SR undefined, so we report successful return to tracking and jerk.
The latter separates a controlled recovery from high-frequency motion that may stand in simulation but is unsuitable for hardware.

\paragraph{Baselines and protocol.}
We compare with three leading real-time general-purpose whole-body trackers: SONIC \citep{luo2025sonic}, HoloMotion-1 \citep{chen2026holomotion}, and Humanoid-GPT \citep{qi2026humanoidgpt}, using their official checkpoints.
For fairness, we restrict baselines to general-purpose whole-body trackers supporting low-latency teleoperation, excluding methods such as PGMT \citep{li2026pgmt} that enable interaction via long future horizons at the expense of real-time responsiveness.
Every policy is evaluated sim-to-sim in MuJoCo \citep{todorov2012mujoco} with the same criteria.
No method receives privileged scene or contact signals.
\shortname{} uses its complete controller, including the filter with $\Rsafe=3$. Hardware deployment runs at $50$\,Hz with a $0.2$\,s look-ahead buffer for ten future reference frames.

\subsection{Generalist Performance Across Regimes}
\label{sec:exp:main}

\begin{table}[t]
  \caption{\textbf{Generalist performance.}
  Sim-to-sim comparison across standard, terrain, implausible-command, and fall benchmarks.}
  \label{tab:main}
  \centering
  \small
  \setlength{\tabcolsep}{3.2pt}
  \begin{adjustbox}{max width=\linewidth}
  \begin{tabular}{lccc ccc cc cc}
    \toprule
    & \multicolumn{3}{c}{\textbf{Standard}} & \multicolumn{3}{c}{\textbf{Terrain}} & \multicolumn{2}{c}{\textbf{Implausible}} & \multicolumn{2}{c}{\textbf{Fall}} \\
    \cmidrule(lr){2-4}\cmidrule(lr){5-7}\cmidrule(lr){8-9}\cmidrule(lr){10-11}
    Method & MPKPE$\downarrow$ & RootVel$\downarrow$ & SR$\uparrow$
           & MPKPE$\downarrow$ & RootPos$\downarrow$ & SR$\uparrow$
           & MPKPE$\downarrow$ & SR$\uparrow$
           & SR$^{\dagger}\uparrow$ & Jerk$\downarrow$ \\
    \midrule
    SONIC \citep{luo2025sonic}              & 82.3  & 189.6          & 94.1          & 331.2         & 294.7          & 15.3          & 327.6          & 50.0          & 5.9           & 1295.5 \\
    HoloMotion-1 \citep{chen2026holomotion} & 109.4 & \textbf{121.3} & 89.0          & 330.0         & 280.7          & 18.7          & 248.7          & 67.6          & 0.7           & 2000.0 \\
    Humanoid-GPT \citep{qi2026humanoidgpt}  & 90.9  & 205.8          & 91.9          & 283.3         & 326.7          & 14.0          & 208.0          & 70.6          & 2.9           & 3598.1 \\
    \midrule
    \shortname{} (ours)                     & \textbf{76.6} & 211.1  & \textbf{96.3} & \textbf{93.3} & \textbf{100.7} & \textbf{81.3} & \textbf{158.0} & \textbf{83.1} & \textbf{99.3} & \textbf{1050.6} \\
    \bottomrule
  \end{tabular}
  \end{adjustbox}
  \smallskip
  \parbox{\linewidth}{\footnotesize \emph{Note:} Position errors are in mm, RootVel in mm/s, Jerk in rad/s$^3$, and SR in \%. $^{\dagger}$Fall SR is recovery rate; all other SR columns treat a fall as failure. Each method is rolled out to the end of every reference.}
\end{table}

While supporting terrain and object interaction and robust fall recovery, \shortname{} maintains strong flat-ground tracking, achieving lower Standard MPKPE (76.6\,mm) and higher success rate (96.3\%) than all three evaluated tracking baselines.
We also evaluate real-robot performance on various demanding motions, including squat jumps, high kicks, and Tai Chi stances.
We also transfer the G1 checkpoint to a Maker L01 humanoid through simple fine-tuning; both sets of results are provided in \cref{app:motion_transfer}.

This advantage becomes more pronounced on Terrain: these baselines reach only 14.0--18.7\% SR and 283.3--331.2\,mm MPKPE, whereas \shortname{} reaches 81.3\% SR and 93.3\,mm MPKPE.
Simply scaling up conventional flat-ground training with randomization therefore does not naturally generalize to terrain interaction.
In contrast, \shortname{} learns the distinct environment-conditioned dynamics across terrains.

Beyond terrain interaction, \shortname{} is also highly robust to both physically implausible commands and falls.
Under physically implausible commands, \shortname{} survives 83.1\% of clips and stays closest to the requested motion, producing a ``best-effort'' response that preserves as much of the original intent as possible without interrupting the task.
The same policy thus falls back gracefully when a command exceeds its capabilities, distinguishing deliberate adaptation to capacity limits from accidental tolerance.
From fallen states, it recovers in 99.3\% of trials without switching controllers, compared with 0.7--5.9\% for the baselines.
It simultaneously achieves the lowest average joint jerk, outperforming all three evaluated baselines on both recovery metrics.

\subsection{Environment Interaction}
\label{sec:exp:interaction}

\begin{figure}[t]
  \centering
  \includegraphics[width=\linewidth]{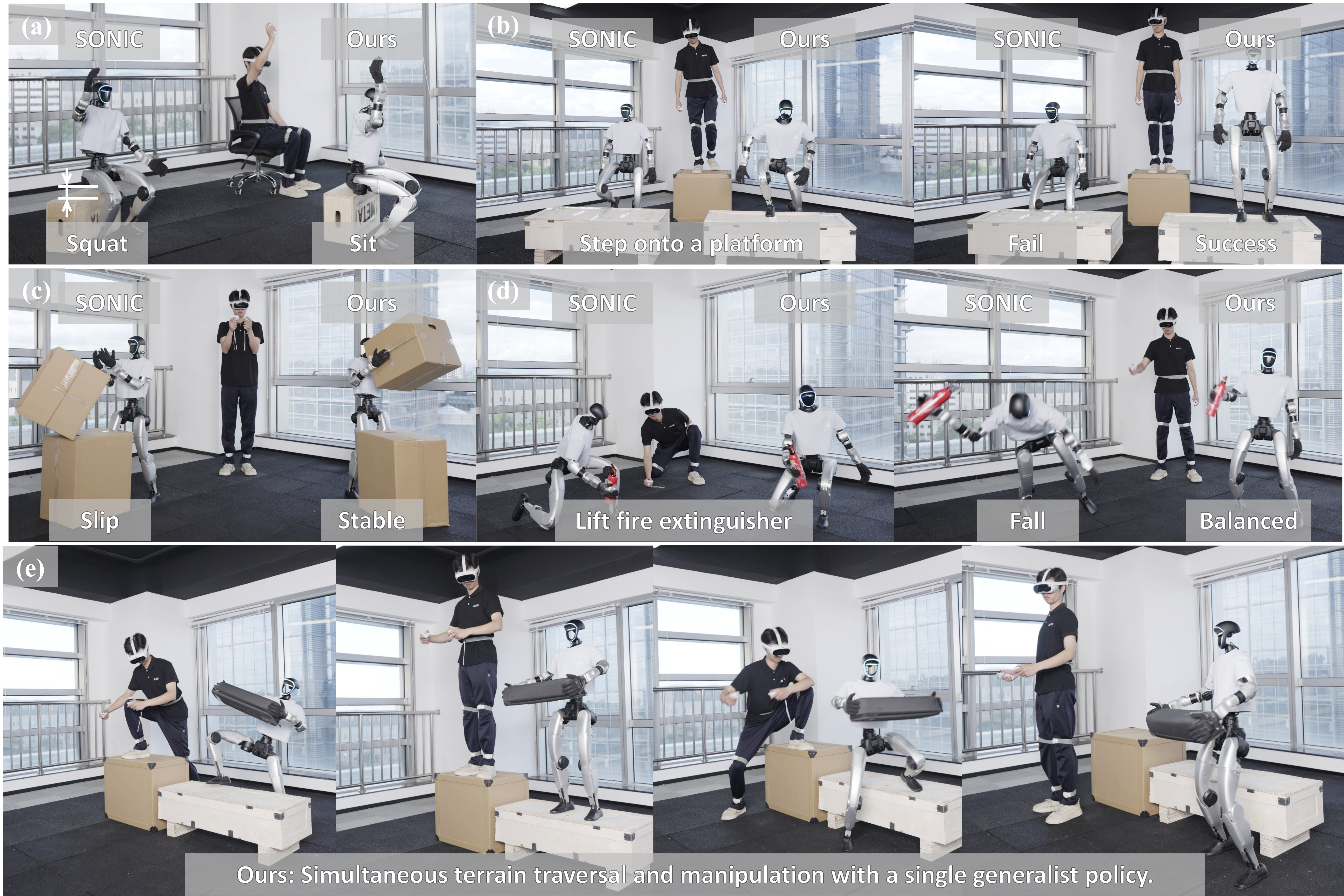}
  \caption{\textbf{Environment interaction on hardware.}
  The same live operator command stream is broadcast simultaneously to the two robots running SONIC (left) and ours (right).
  \textbf{(a,b)} Terrain interaction: sitting on a box and stepping onto a platform.
  \textbf{(c,d)} Object interaction: lifting a box and rising with a fire extinguisher.
  \textbf{(e)} Unified interaction: ours carries a case while stepping onto and down from a platform.}
  \label{fig:interaction}
\end{figure}

The matched hardware trials in \cref{fig:interaction}(a,b) compare terrain interaction with the same live operator command stream broadcast simultaneously to both robots.
SONIC fails to exploit the terrain for meaningful interaction: it only half-squats without sitting and raises a leg without using the platform to step up, whereas \shortname{} completes both motions.
Panels (c,d) compare object interaction.
In (c), \shortname{} applies the forces required to lift and hold the box, whereas SONIC merely imitates the lifting motion and drops it.
In (d), \shortname{} captures and adapts to the payload-induced dynamics of the fire extinguisher, lifting it stably, whereas SONIC loses balance and falls.
Panel (e) further shows that the same policy can simultaneously perform cross-terrain mobility and object manipulation: it carries a case across a platform in one continuous sequence without switching controllers.

\subsection{Robustness}
\label{sec:exp:robust}

\begin{figure}[t]
  \centering
  \includegraphics[width=\linewidth]{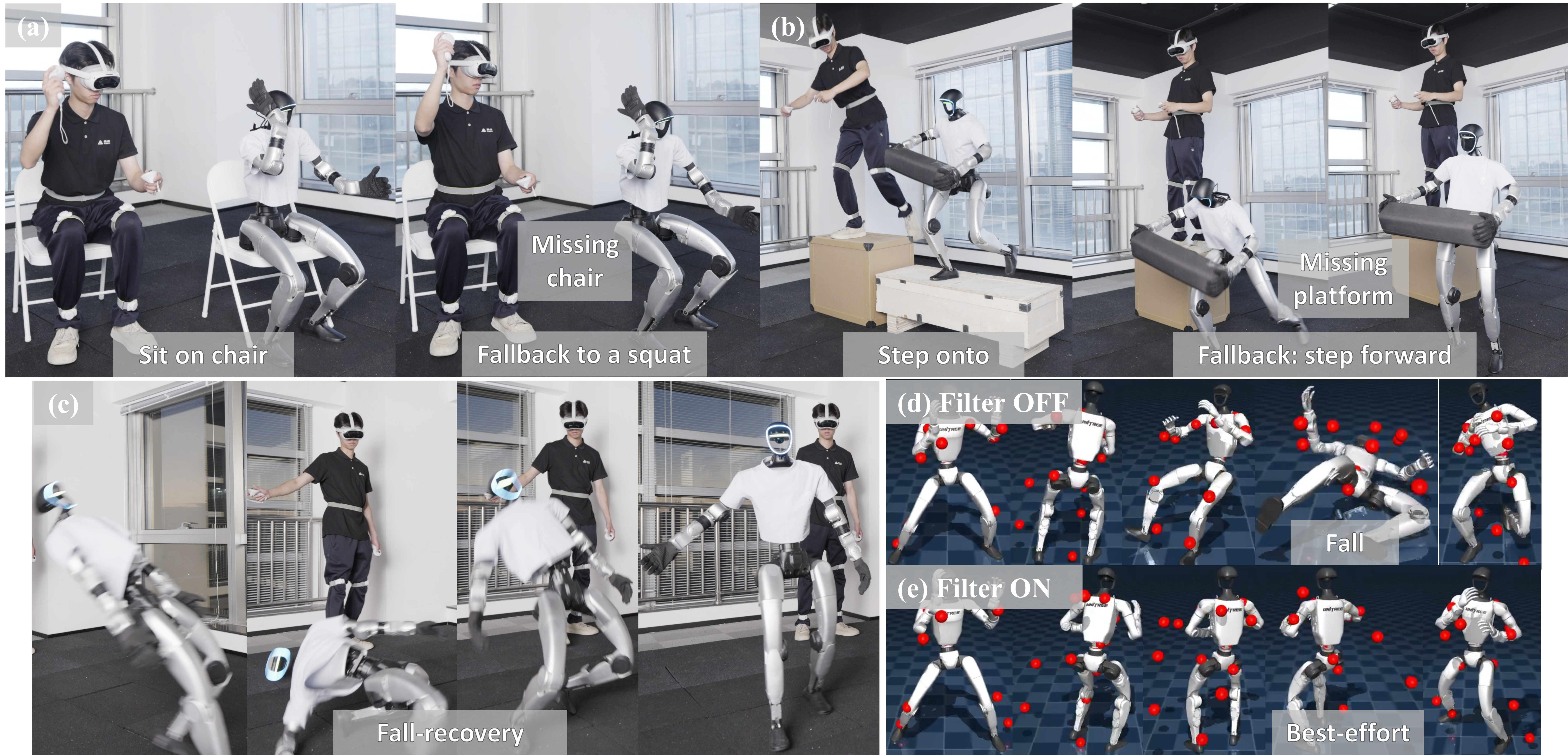}
  \caption{\textbf{Robustness to invalid supports, disturbances, and commands.}
  \textbf{(a,b)} With an expected chair or platform absent, the policy settles into a stable best-effort motion.
  \textbf{(c)} The tracking policy recovers after a kick-induced fall.
  \textbf{(d,e)} For the same physically implausible spinning-kick reference, filtering suppresses the destabilizing portion while retaining the commanded turn.}
  \label{fig:robustness}
\end{figure}

The hardware sequences in \cref{fig:robustness}(a,b) remove supports assumed by the command.
Rather than commit to nonexistent contact or freeze, the policy preserves the achievable part of the motion.
The same tracker stands up after being knocked down in panel (c).
Panels (d,e) present a qualitative ablation of the command filter in simulation: without it, the controller attempts a self-interfering kick and falls (though later gets back up autonomously); with it, the controller makes a safe best-effort response---turning as commanded while suppressing the destabilizing kick.

\subsection{Ablation Study}
\label{sec:exp:ablation}

\begin{table}[t]
  \caption{\textbf{Component ablations.}
  Effects of the BWM, terrain data, and robustness mechanisms across all benchmarks.}
  \label{tab:ablation}
  \centering
  \small
  \setlength{\tabcolsep}{1.8pt}
  \begin{adjustbox}{max width=\linewidth}
  \begin{tabular}{lccc ccc ccc cc cc}
    \toprule
    & \multicolumn{3}{c}{\textbf{Components}}
    & \multicolumn{3}{c}{\textbf{Standard}}
    & \multicolumn{3}{c}{\textbf{Terrain}}
    & \multicolumn{2}{c}{\textbf{Implausible}}
    & \multicolumn{2}{c}{\textbf{Fall}} \\
    \cmidrule(lr){2-4}\cmidrule(lr){5-7}\cmidrule(lr){8-10}\cmidrule(lr){11-12}\cmidrule(lr){13-14}
    Variant & \makecell{BWM\\(\S~\ref{sec:bwm})} & \makecell{Terrain\\(\S~\ref{sec:terrain})} & \makecell{Robust.\\(\S~\ref{sec:filter})}
            & MPKPE$\downarrow$ & RootVel$\downarrow$ & SR$\uparrow$
            & MPKPE$\downarrow$ & RootPos$\downarrow$ & SR$\uparrow$
            & MPKPE$\downarrow$ & SR$\uparrow$
            & SR$^{\dagger}\uparrow$ & Jerk$\downarrow$ \\
    \midrule
    w/o \S~\ref{sec:bwm} BWM          & \xmark & \cmark & --     & 130.3 & \textbf{184.3} & 80.1 & 102.9 & 120.4 & 76.7 & 236.3 & 64.7 & 9.6 & \textbf{295.3} \\
    w/o \S~\ref{sec:terrain} Terrain Data & \cmark & \xmark & \cmark & \textbf{75.2} & 210.5 & \textbf{97.8} & 220.9 & 311.7 & 12.0 & \textbf{156.2} & \textbf{83.8} & \textbf{100.0} & 1033.7 \\
    \midrule
    Complete \shortname{}     & \cmark & \cmark & \cmark & \textbf{76.6} & 211.1 & \textbf{96.3} & \textbf{93.3} & \textbf{100.7} & \textbf{81.3} & \textbf{158.0} & \textbf{83.1} & \textbf{99.3} & 1050.6 \\
    \bottomrule
  \end{tabular}
  \end{adjustbox}
  \smallskip
  \parbox{\linewidth}{\footnotesize \emph{Note:} Each component column represents the module described in the corresponding method subsection; ``--'' denotes joint removal of the robustness module with the BWM, reflecting the filter's dependence on BWM predictions. Bold marks the best and near-best values relative to the overall spread.}
\end{table}

\Cref{tab:ablation} treats the contents of \cref{sec:bwm,sec:terrain,sec:filter} as the three components.
\emph{w/o \S~\ref{sec:bwm} BWM} removes the core behavior-world-model mechanism in \cref{sec:bwm}: future predictions and their associated objectives.
It also removes the robustness module in \cref{sec:filter}, including the prediction-dependent command filter and fallen-state initialization.
\emph{w/o \S~\ref{sec:terrain} Terrain Data} retains the BWM and robustness mechanisms but removes the annotation module in \cref{sec:terrain}: no terrain is generated for training.
All evaluation metrics follow the definitions in \cref{tab:main}.

Jointly ablating the BWM and robustness mechanisms (\cref{sec:bwm,sec:filter}) reduces recovery from 99.3\% to 9.6\% and raises MPKPE across all three regimes, demonstrating the integrated approach's contribution.
The recovery loss is consistent with fallen-state initialization exposing the policy to recovery scenarios, while the pronounced reduction in Implausible SR is consistent with the command filter suppressing implausible commands.
The increased MPKPE suggests a benefit from predictive supervision, with next-state prediction providing a plausible mechanism through learning interaction dynamics.
Although this ablation does not isolate individual contributions, its results support the proposed component-level interpretations.

Removing Terrain Data substantially degrades every metric on Terrain, with SR dropping from 81.3\% to 12.0\% and MPKPE increasing from 93.3\,mm to 220.9\,mm. This result shows that incorporating our automatically annotated terrain data into training markedly improves terrain interaction performance.
A quantitative sensitivity analysis of the filter radius is provided in Appendix~\ref{app:tradeoff}; the spinning-kick comparison in \cref{sec:exp:robust} provides a qualitative filter ablation, and, for brevity, additional relevant experiments are presented in Appendix~\ref{app:motion_transfer}.

%% file: sections/conclusion.tex
\section{Conclusion}
\label{sec:conclusion}

We introduced \shortname{}, a behavior world model for highly robust whole-body tracking with environment interaction.
Jointly predicting actions, future states, and future behaviors lets the policy adapt to contact-conditioned dynamics and produce stable ``best-effort'' behavior under implausible commands.
An automatic spatial terrain annotation pipeline supplies paired motion--terrain data at scale for training.
\shortname{} combines terrain and object interaction and robust fall recovery with strong flat-ground tracking, achieving lower Standard MPKPE (76.6\,mm) and higher success rate (96.3\%) than all three evaluated tracking baselines.
It reaches 81.3\% success on terrain interaction ($4.3\times$ the best real-time general-purpose baseline) and 83.1\% survival under physically implausible commands, while raising fall recovery rate from single-digit baseline rates to 99.3\%.
Hardware experiments further demonstrate agile teleoperation, terrain and object interaction, and recovery from disturbances.
These capabilities share one policy and a live command interface.
Two limitations remain: the command filter recognizes unfamiliar behavior rather than physical risk, and the contact-based annotation is suited to static terrain but not finite-mass movable objects and reconstructs touched supports rather than complete scenes.
Extending the reconstruction to non-supporting geometry, and connecting the world model's state prediction to a more direct notion of physical risk, are natural next steps.

%% file: sections/appendix.tex
\section{Policy Interface and Model Architecture}
\label{app:input}

The controller operates the 29 actuated degrees of freedom of a Unitree G1 at $50$\,Hz.
Its proprioceptive input $\prop_t\in\mathbb{R}^{67}$ contains projected gravity, base angular velocity, torso linear acceleration, and joint positions and velocities.
The policy also receives the previous PD target $\act_{t-1}\in\mathbb{R}^{29}$ and outputs the next target $\act_t\in\mathbb{R}^{29}$.
The reference window contains ten future frames sampled at $0.02$\,s intervals. During live teleoperation, buffering the incoming operator stream introduces a fixed $0.2$\,s look-ahead delay. The policy has been verified to run at $50$\,Hz on hardware.
At each frame, it contains 29 joint positions, the 6D reference-root rotation relative to the current robot root, frame-to-frame root translation, reference-root translation relative to the robot, and gravity expressed in the reference-root frame, giving $\refwin_t\in\mathbb{R}^{440}$.

The FSQ reconstruction branch uses 32 levels per latent dimension.
The core input $\bm e_t=[\prop_t,\act_{t-1},\lat_t]\allowbreak\in\mathbb{R}^{160}$ is projected to width 256, and the Transformer uses four attention heads.

\section{Implementation Details}
\label{app:training}

We optimize with PPO \citep{schulman2017proximal} in Isaac Lab \citep{mittal2025isaaclab} over the corpus mix described in \cref{sec:terrain}.
Training samples terrain-annotated motion with probability 0.2 and flat-ground motion otherwise.
In \cref{eq:objective}, $\lambda_{\mathrm{recon}}=0.01$, $\lambda_{\mathrm{cycle}}=1$, $\lambda_{\mathrm{state}}=0.02$, and $\lambda_{\mathrm{beh}}=0.01$; next-step terms are masked across episode boundaries.
Each sequence-level update recomputes the current rollout segment with gradients while attaching the detached per-layer KV prefix saved at rollout start, so the attention context during optimization matches the context under which actions were sampled.

Command robustness comes from the filter in \cref{sec:filter}, whereas fall robustness is trained directly within the tracker by starting a fraction of flat-ground episodes from synthetic fallen poses under a curriculum over trunk inclination up to fully prone.
A validated whole-body collision proxy places the robot at ground contact, and precise tracking rewards are gated by a smooth target-relative recovery gate so the policy is not penalized too hard while standing up.
Separately, persistent random external forces at the wrists and torso emulate payloads and contact loads.

\begin{table}[ht]
  \caption{Reward terms (left) and domain randomization (right).
  Tracking terms use the Gaussian kernel in \cref{eq:reward}; other terms are direct penalties.}
  \label{tab:impl}
  \centering
  \footnotesize
  \begin{minipage}[t]{0.47\linewidth}
    \centering
    \setlength{\tabcolsep}{2.8pt}
    \begin{tabular}{lrr}
      \toprule
      Reward term & Weight & $\sigma$ \\
      \midrule
      Anchor position            & $1.0$  & $0.30$ \\
      Anchor orientation         & $0.5$  & $0.40$ \\
      Relative body position     & $1.0$  & $0.30$ \\
      Relative body orientation  & $1.0$  & $0.40$ \\
      Body linear velocity       & $1.0$  & $1.00$ \\
      Body angular velocity      & $1.0$  & $3.14$ \\
      Local head/wrist points    & $2.0$  & $0.10$ \\
      Local wrist orientation    & $2.0$  & $0.30$ \\
      Local feet position        & $1.0$  & $0.15$ \\
      \midrule
      Action rate                & $-1\mathrm{e}{-1}$   & -- \\
      Joint limit                & $-10.0$              & -- \\
      Undesired contacts         & $-1\mathrm{e}{-1}$   & -- \\
      Anti-shake angular vel.    & $-5\mathrm{e}{-3}$   & -- \\
      Ankle joint acceleration   & $-2.5\mathrm{e}{-6}$ & -- \\
      \bottomrule
    \end{tabular}
  \end{minipage}\hfill
  \begin{minipage}[t]{0.51\linewidth}
    \centering
    \setlength{\tabcolsep}{2.7pt}
    \begin{tabular}{llr}
      \toprule
      Randomized quantity & Range & Resample \\
      \midrule
      Static friction        & $[0.3,1.6]$          & startup \\
      Dynamic friction       & $[0.3,1.2]$          & startup \\
      Restitution            & $[0.0,0.5]$          & startup \\
      Wrist/torso mass       & $\times[0.8,2.5]$    & startup \\
      Torso CoM $x$          & $\pm0.025$\,m        & startup \\
      Torso CoM $y,z$        & $\pm0.05$\,m         & startup \\
      Joint default offset   & $\pm0.01$\,rad       & startup \\
      \midrule
      Push linear $x,y$      & $\pm0.5$\,m/s        & 4--6\,s \\
      Push linear $z$        & $\pm0.2$\,m/s        & 4--6\,s \\
      Push angular $r,p$     & $\pm0.52$\,rad/s     & 4--6\,s \\
      Push angular yaw       & $\pm0.78$\,rad/s     & 4--6\,s \\
      Wrist/torso force      & $\pm5$\,N            & 2--5\,s \\
      Wrist/torso torque     & $\pm0.3$\,N\,m       & 2--5\,s \\
      Root-translation mask  & masked w.p.\ $2/3$   & 2--5\,s \\
      Reference-window noise & $\pm0.05$            & each step \\
      \bottomrule
    \end{tabular}
  \end{minipage}
\end{table}

Each tracking term maps an error $\bm e$ to
\begin{equation}
r=\exp\!\left(-\lVert\bm e\rVert^2/\sigma^2\right).
\label{eq:reward}
\end{equation}
Table~\ref{tab:impl} lists the weights, kernel widths, and randomization ranges.
Anchor terms act on the root, relative-body terms on tracked links in the root frame, and local terms on the head proxy, wrists, and feet in the root frame, matching the points available to a VR operator.
The critic is discarded at deployment and uses a separate MLP over 1756-dimensional privileged observations: multi-frame reference targets, root pose differences, body poses, and ten frames of proprioception and action history.

For Standard, Terrain, and Implausible, success rate (SR) uses a deviation-from-reference criterion: a rollout fails if the translation or orientation of either the base or torso link deviates from the reference motion beyond the corresponding threshold, or if the robot falls.
For Fall, a rollout succeeds once none of these failure conditions holds before the motion sequence ends.

\section{Real-Robot Motion Tracking and Cross-Embodiment Transfer}
\label{app:motion_transfer}

\begin{figure}[ht]
  \centering
  \includegraphics[width=\linewidth]{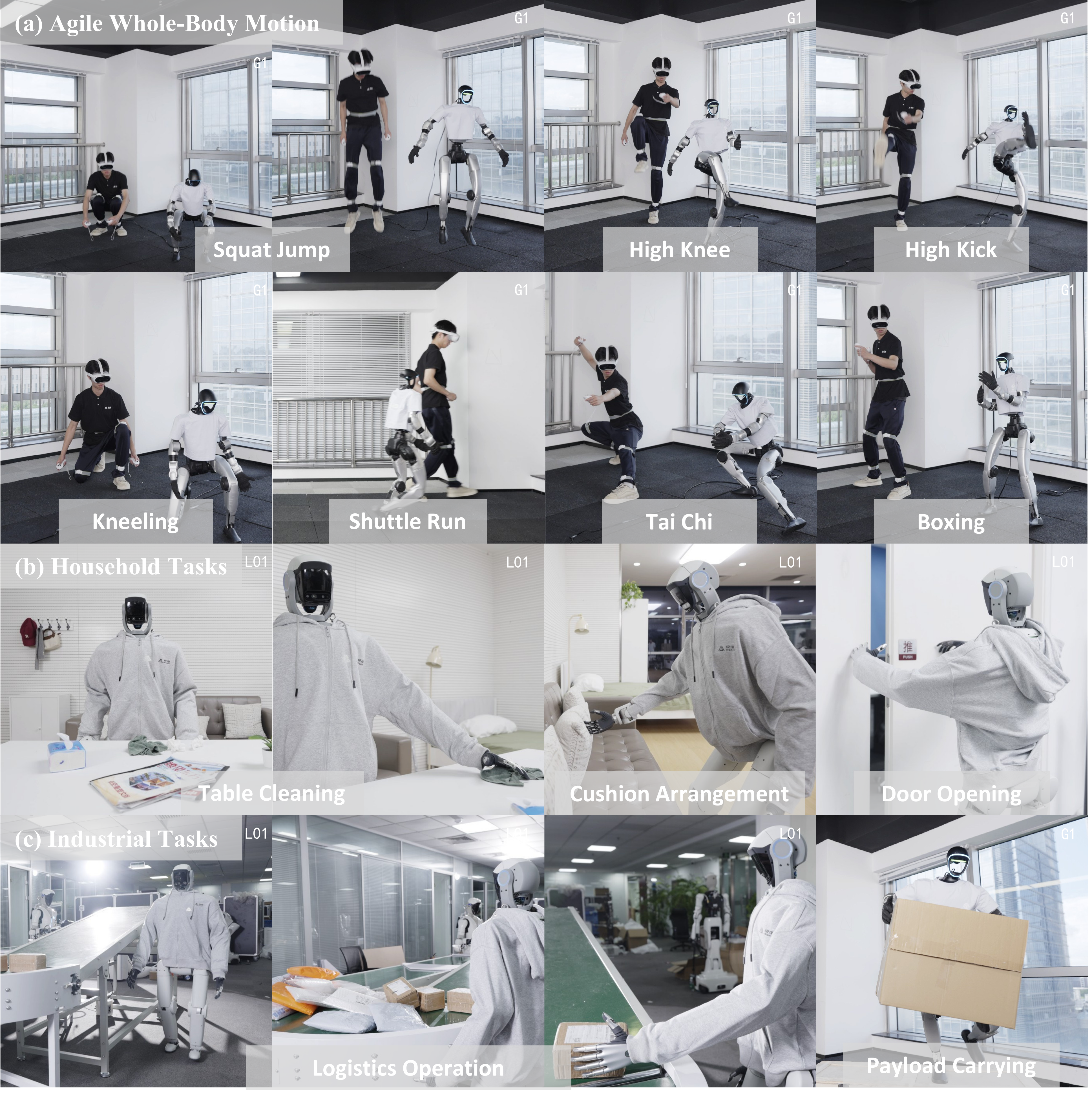}
  \caption{\textbf{Whole-body motion capability.}
  Agile motion, household tasks, and industrial tasks under live whole-body command use the same policy architecture and training approach.
  Frames identify the robot: Unitree G1 uses the original checkpoint, while Maker L01 uses a checkpoint fine-tuned from it.}
  \label{fig:motion_capability}
\end{figure}

\Cref{fig:motion_capability} collects real-robot results beyond the primary claims evaluated in the main paper.
With the same live whole-body command interface, the policy tracks demanding flat-ground motions including squat jumps, high kicks, and low Tai Chi stances.
These trials demonstrate that \shortname{} retains agile flat-ground tracking alongside its terrain and object interaction and robust fall-recovery capabilities, achieving lower Standard MPKPE (76.6\,mm) and higher success rate (96.3\%) than all three evaluated tracking baselines (\cref{tab:main}).

We additionally test whether the learned controller can be transferred across embodiments.
We retarget the motion corpus to the Maker L01 humanoid and fine-tune the Unitree G1 checkpoint with the same training recipe.
The resulting policy recovers whole-body tracking on the new robot and drives household and industrial behaviors under live command, coordinating locomotion with object interaction without switching to a task-specific manipulation stack.
The Maker L01 household and logistics trials in \cref{fig:motion_capability} not only demonstrate cross-embodiment transfer qualitatively, but also show that the policy can support remote teleoperation and data collection in household and industrial settings, providing embodied data for future training of upstream ``brain'' models.

\section{Safety-Radius Sensitivity}
\label{app:tradeoff}

\begin{figure}[ht]
  \centering
  \includegraphics[width=0.58\linewidth]{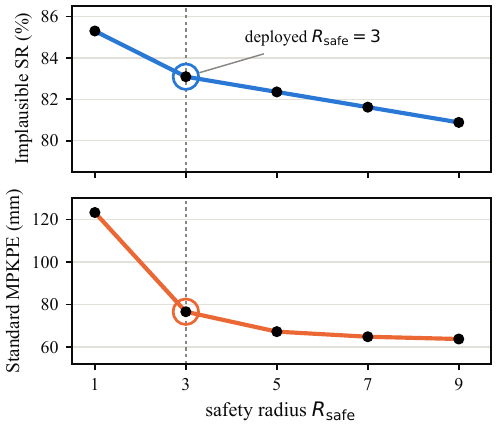}
  \caption{Standard-split MPKPE and Implausible SR as a function of the safety radius.
  $\Rsafe=3$ is used in the main experiments.}
  \label{fig:tradeoff}
\end{figure}

We use $\Rsafe=3$ throughout as a robustness-oriented trade-off.
Tightening the radius to $\Rsafe=1$ increases Standard MPKPE by 46.7\,mm for just 2.2 more Implausible SR percentage points.
For $\Rsafe>3$, Standard MPKPE is comparatively flat, whereas Implausible SR remains visibly responsive to the radius.

\section{Human Audit of Terrain Annotations}
\label{app:terrain_audit}

The automatic pipeline in \cref{sec:terrain} generates annotations without supervision, so no ground-truth terrain exists for direct comparison.
Geometric failures such as body--terrain penetration are already filtered during generation.
What remains requires human judgment: inferring the intended interaction, assessing whether its implied contacts are physically sound, and deciding whether motion and terrain together look like something a person would actually do.

\begin{table}[ht]
  \centering
  \caption{Audit of terrain annotations and dominant failure modes.}
  \label{tab:terrain_acc}
  \small
  \setlength{\tabcolsep}{4pt}
  \begin{tabularx}{\linewidth}{@{}lcc>{\raggedright\arraybackslash}X@{}}
    \toprule
    Terrain category & \# sampled & Correct (\%) & Dominant failure mode \\
    \midrule
    Stairs / steps & 50  & 92 & Fast footfall obscures the deceleration signature \\
    Chair / seat   & 50  & 94 & Torso sway misplaces the backrest contact \\
    Box / platform & 50  & 98 & Sparse contacts fail to form a primitive \\
    Other          & 50  & 84 & Hand-retargeting noise hides the velocity drop \\
    \midrule
    Overall        & 200 & 92 & --- \\
    \bottomrule
  \end{tabularx}
\end{table}

We audit 200 randomly sampled Bones-Seed annotations, 50 in each category of \cref{tab:terrain_acc}.
An annotation is judged correct only when its reconstructed supports match the intended interaction and make the paired motion--terrain behavior plausible.
Bones-Seed is used because its cleaner retargeting better isolates annotation error.
Overall accuracy is 92\%.
Platforms are most reliable; missed short contacts account for their rare failures.
Fast footfalls or torso sway can displace stair and seat supports, while wrist retargeting noise makes the hand-contact-heavy ``Other'' category hardest.
As expected, the annotation is conservative: errors remove or misplace a support rather than introduce an intersecting obstacle.